\documentclass[twoside,leqno,twocolumn,10pt]{article}
\usepackage[letterpaper]{geometry}

\usepackage{MFCS2020/ltexpprt}
\usepackage{paralist}

\usepackage{amssymb,amsmath,amsthm}

\usepackage{algorithm}
\usepackage{algpseudocode}

\newcommand{\X}{\mathbf{X}}
\newcommand{\V}{\mathbf{V}}
\usepackage{times}
\usepackage{microtype}
\usepackage{graphicx}
\usepackage{paralist}
\usepackage{float}
\usepackage{todonotes}
 \allowdisplaybreaks

\usepackage{paralist}

\newcommand\numberthis{\addtocounter{equation}{1}\tag{\theequation}}
\usepackage{hyperref}
\usepackage{amsfonts}
\usepackage{amssymb}
\usepackage{color}
\newcommand{\Tr}{\mathrm{Tr}}

\newcommand{\E}{\mathbb{E}}

\newcommand{\F}{\mathbf{F}}
\newcommand{\G}{\mathbf{G}}

\usepackage{bbm}
\usepackage{verbatim}

\newcommand{\Var}{\mathrm{Var}}

\newcommand{\R}{\mathbb{R}}

\newcommand{\Cov}{\mathrm{Cov}}

\newtheorem{thm}{Theorem}
\newtheorem{newfact}[thm]{Fact}
 \newtheorem{defi}[thm]{Definition}
 \newtheorem{lem}[thm]{Lemma}
  \newtheorem{cor}[thm]{Corollary}
  \newtheorem{remark}[thm]{Remark}

\newcommand*{\email}[1]{\href{mailto:#1}{\nolinkurl{#1}} } 
\begin{document}

\title{\Large Improving \textit{Tug-of-War} sketch using Control-Variates method}
\author{Rameshwar Pratap\thanks{IIT Mandi, H.P., India (\email{rameshwar.pratap@gmail.com}).}
\and Bhisham Dev Verma\thanks{IIT Mandi, H.P., India (\email{bhishamdevverma@gmail.com}).}
\and Raghav Kulkarni\thanks{Chennai Mathematical Institute, Chennai, India (\email{kulraghav@gmail.com}).}}
\date{}

\maketitle


\begin{abstract}
Computing space-efficient summary, or \textit{a.k.a. sketches}, of large data, is a central problem in the streaming algorithm. Such sketches are used to answer \textit{post-hoc} queries in several data analytics tasks. The algorithm for computing sketches typically requires to be fast, accurate, and space-efficient. A fundamental problem in the streaming algorithm framework is that of computing the frequency moments of data streams. The frequency moments of  a sequence containing $f_i$ elements of type $i$, are the numbers $\F_k=\sum_{i=1}^n {f_i}^k,$  where $i\in [n]$. 
This is also called as  $\ell_k$ norm of the frequency vector $(f_1, f_2, \ldots f_n).$ 
Another important problem is to compute the similarity between two data streams by computing the inner product of the corresponding frequency vectors. The seminal work of Alon, Matias, and Szegedy~\cite{AMS}, \textit{a.k.a. Tug-of-war} (or AMS) sketch gives a randomized sublinear space (and linear time) algorithm for computing the frequency moments, and the inner product between two frequency vectors corresponding to the  data streams. However, the variance of these estimates typically tends to be large. In this work, we focus on  minimizing  the variance of these estimates.  We use the techniques from the classical Control-Variate method~\cite{Lavenberg} which is primarily known for variance reduction in Monte-Carlo simulations, and as a result, we are able to obtain 
significant variance reduction, at the cost of a little computational overhead.  We present a theoretical analysis of our proposal and complement it with supporting experiments on synthetic as well as real-world datasets.


\end{abstract}
\section{Introduction}
Several data analytics tasks require analysing massive data stream such as real-time IP traffic analysis~\cite{EstanV02}, metagenomics~\cite{gkaa265},  email/tweets/SMS, time-series data~\cite{ZhuS02}, web clicks and crawls, sensor/IoT readings~\cite{inproceedings}. In many of these applications we may not have enough space/memory available to store the entire data stream.  Moreover, these  data packets arrive at a rapid rate. Therefore analysing such a large data stream requires sophisticated algorithmic techniques   that can not only maintain a small footprint of the datastream but also incorporate the fast dynamic updates. Such a small footprint is commonly known as a \textit{sketch} of the data stream, and is useful in answering important properties about the data stream during post-hoc queries~\cite{Muthukrishnan05}.


In this paper, we consider the problems of estimating: \begin{inparaenum}[i)]
\item  frequency moments and of the data stream, and 
\item  inner product of frequency vectors corresponding to a pair of data streams. 
\end{inparaenum}
 Let $\sigma_1=\{a_1, a_2, \ldots, a_{m_1} \}$ and $\sigma_2=\{b_1, b_2, \ldots, b_{m_2} \}$ 
be two data streams of lengths $m_1$ and $m_2$, respectively, where  $\forall i\in[m_1], j\in[m_2]$, we have $a_i, b_j\in[n]$. This also implicitly defines the frequency vector over the elements in the data stream.  We denote $\mathbf{f}=\langle f_1, f_2, \ldots, f_n \rangle$, and $\mathbf{g}=\langle g_1, g_2, \ldots, g_n \rangle$, where $f_i, g_i$ denotes the number of occurrences of the element $a_i, b_i$ in the stream $\sigma_1$ and $\sigma_2$, respectively . We denote  $k$-frequency moments (or $\ell_k$ norm) of the frequency vector by $\F_k$ and define it as follows: 
\[
\F_k=\sum_{i=1}^n f_i^k. 
\]
\footnote{Note that $\F_0$ corresponds to the number of distinct elements in the stream $n$, and $\F_1$ corresponds to the number of elements in the stream $m$.
}
In this work, we focus on the case when $k= 2$. We define the inner product of the frequency vectors $\mathbf{f}$ and  $\mathbf{g}$ as follows:
\[
\langle \mathbf{f}, \mathbf{g} \rangle=\sum_{i=1}^n{f_i}{g_i}.
\]
The seminal work of Alon, Matias and Szegedy~\cite{AMS} gives a sublinear space algorithm, 
for estimating the frequency moments, and the inner product between the frequency vectors. However, the variances of their estimates tend to be large when the data streams have many items of large frequencies.   In this work, we address this challenge and suggest a technique that can reduce the variances provided by the AMS-sketch, and as a consequence give a more accurate estimation. We use the classical control variate method for the purpose, and we discuss it as follows:

\subsection{ Our approach -- variance reduction using control variate trick:} \label{subsec:CV_trick}
 
Our technique exploits the control-variate trick used for reducing variance that occurs while estimating the frequency moments via AMS-sketch~\cite{AMS} for the large data streams. The control-variate is a classical method that is being used for variance reduction in Monte-Carlo simulation, which is done via analyzing the correlated errors~\cite{Lavenberg}. We illustrate this with an example as follows: suppose we have a process generating a random variable $X$, and we are interested in estimating the quality of $\E[X]$. Let us have another process for generating another random variable $Z$ such that we know exactly the value of its true mean $\E[Z]$. Then for any constant $c$, the expression $X+c(Z-\E[Z])$ is an unbiased estimator of $X$, due to the following:

\begin{align*}
    \E[X+c(Z-\E[Z])]&=\E[X]+c\E[Z-\E[Z]].\\&=\E[X]+0=\E[X]\numberthis \label{eq:cv_exp}.
\end{align*}
The variance of the expression $X+c(Z-\E[Z])$ is given by 
\begin{align*}
    \Var[X+c(Z-\E[Z])]&=\Var[X]+c^2\Var[Z]+2c \Cov[X,Z]\numberthis \label{eq:CV_var}.
\end{align*}

By elementary calculus we can find the appropriate value of $c$ which minimise the above expression. Suppose we denote that value by $\hat{c}$, then
\[
\hat{c}=-\frac{\Cov[X,Z]}{\Var[Z]} \numberthis \label{eq:c_hat}.
\]
Equations~\ref{eq:CV_var}, \ref{eq:c_hat} give us the following
\[
 \Var[X+c(Z-\E[Z])]=\Var[X]-\frac{\Cov[X,Z]^2}{\Var[Z]}\numberthis \label{eq:cv_correction}.
\]
To summarize the above, for a random variable $X$, we are able to generate another random variable $X+c(Z-\E[Z])$ such that both have the same expected value, i.e.,  the random variable $X+c(Z-\E[Z])$ is an unbiased estimator of $X$. Further, the variance of $X+c(Z-\E[Z])$ is smaller than or equal to that of $X$ because the ${\Cov[X, Z]^2}/{\Var[Z]}$ is always non-negative --  with the equality if there is no correlation between $X$ and $Z$. The random variable $Z$ is called control variate, and the term $\hat{c}$ is called the control variate coefficient. In order to apply control variate trick for variance reduction, there are some practical considerations that need to be addressed carefully such as choosing an appropriate random variable $Z$, computing its expected value, and then choosing the coefficient $c$, etc.

\subsection{Our results:}
As an application of the control variate trick, we provide significant variance reduction in the frequency moments estimators of AMS-sketch.  We present our theoretical guarantees on the variance reduction for $\F_2$ as follows.



\begin{thm}\label{thm:AMS}
Let $X$ be the random variable denoting the estimate of $\F_2$  
in AMS-Sketch~\cite{AMS} (see Algorithm~\ref{alg:AMS}, Theorem~\ref{thm:AMS_vanilla}). 
Then
there exists a control variate random variable $Z$, and the corresponding control variate coefficient $\hat{c}$ such that $$\Var(X + \hat{c}(Z - \E[Z])) = \Var(X)- \frac{\left(\F_1^2-\F_2\right)^2}{\F_0(\F_0-1)},$$
where, $\E[X]=\F_2\text{~and,~}\Var(X) = 2(\F_2^2-\F_4),$ as noted in ~\cite{AMS} (Theorem~\ref{thm:AMS_vanilla}). 
\end{thm}

 We further extend our results which is used for  variance reduction in estimating the  inner product of two frequency vectors corresponding to a pair of  data streams. 
We present our results as follows:

\begin{thm}\label{thm:AMS_IP_estimation}
 Let $\Tilde{f}$ and $\Tilde{g}$ be the sketches of $\mathbf{f}$ and $\mathbf{g}$ obtained via Tug-of-War sketch, respectively. Let $X^{(2)} $ be the random variable denoting the estimate of $\langle \mathbf{f}, \mathbf{g} \rangle$  
in AMS-Sketch~\cite{AMS} (see Algorithm~\ref{alg:AMS}, Theorem~\ref{thm:AMS_vanilla_dot_product}). 
Then
there exists a control variate random variable $Z^{(2)}$, and the corresponding control variate coefficient $\hat{c}$ such that
\begin{align*}
 &\Var(X^{(2)} + \hat{c}(Z^{(2)} - \E[Z^{(2)}])) \\&= \Var(X^{(2)})- \frac{2\left(\langle \mathbf{f}, \mathbf{g} \rangle(\F_2+\G_2)\right)^2}{2\langle \mathbf{f}, \mathbf{g} \rangle^2+{\F_2}^2+{\G_2}^2},
 \end{align*}
where, $\E[X^{(2)}]=\langle \mathbf{f}, \mathbf{g} \rangle, \text{~and,~}\Var(X^{(2)}) = \sum_{i\neq j}{f_i}^2{g_j}^2+\sum_{i\neq j, i, j \in [n]} f_ig_if_jg_j,$ as noted in ~\cite{AMS} (Theorem~\ref{thm:AMS_vanilla_dot_product}). \end{thm}

\subsubsection{Comment on  the overhead of our estimate:} 
\noindent\subsubsection*{For estimation of $\F_2$ -- Theorem~\ref{thm:AMS}:} Our control variate  random variable $Z$ is independent of the actual values of the stream and we only need to know the number of distinct elements, $n$, in the stream in order to construct $Z.$
Therefore,  our new random variable $X + \hat{c}(Z-\E[Z])$ can be estimated with a small computational overhead, that is,  $O(\log {n})$ space and $O(n)$ time, which is additive to the AMS algorithm.

 The mean of the control variate random variable $Z$ (see Equations~\eqref{eq:CV_AMS},\eqref{eq:exp_z_AMS}) is  zero. As we can exactly compute the true mean of $Z$, we have  $\E[X + c(Z-\E[Z])] = \E[X].$ Therefore, our new  estimate does not introduce any additional bias to the respective estimate of AMS sketch.
 
 The optimum value of $c$, given by $\hat{c}$ in Equation~\eqref{eq:control_variate_CMS}
 turns out to be  
\[
\hat{c}=-\frac{\Cov[X,Z]}{\Var[Z]}=-\frac{\F_1^2-\F_2}{\F_0(\F_0-1)}. 
\]
In practice, we choose an approximation for $\hat{c}$.
 The $\mathbf{F}_1$ is the length of the stream $\sigma$ and its value is  $m$, which we assume is known to us. We also assume that we know the value of $\mathbf{F}_0$.\footnote{We do not need to know the exact value of $\mathbf{F}_0$, knowing an approximate upper bound would suffice.}
We take the approximation of $\mathbf{F}_2$ obtained using the AMS sketch as a proxy for $\F_2$.

\noindent\subsubsection*{For estimation of inner product -- Theorem~\ref{thm:AMS_IP_estimation}:} Our control variate  random variable $Z^{(2)}$ is the estimate of  sum of $\ell_2$ norm of frequency vectors $\mathbf{f}$ and $\mathbf{g}$ (see Equation~\eqref{eq:cv_dot_product_defintion}). The mean of the control variate random variable $Z^{(2)}$ (see Equation~\eqref{eq:cv_dot_product}) is  $\F_2+\G_2$.  The optimum value of $c$, given by $\hat{c}$ in Equation~\eqref{eq:control_variate_CMS}
 is 
\[
\hat{c}=-\frac{\langle \mathbf{f}, \mathbf{g} \rangle(\F_2+\G_2)}{(2\langle \mathbf{f}, \mathbf{g} \rangle^2+{\F_2}^2+{\G_2}^2)}. 
\]
In practice, we choose  approximations for $\hat{c}$ and $\E[Z^{(2)}]$. 
We take the approximation of $\mathbf{F}_2, \G_2$ and $\langle \mathbf{f}, \mathbf{g} \rangle$ obtained using the AMS sketch as  proxies of the corresponding values.  Therefore,  our new random variable $X^{(2)} + \hat{c}(Z^{(2)}-\E[Z^{(2)}])$ can be estimated with a small computational overhead, that is,  $O(\log {n})$ space and $O(n)$ time, to the AMS algorithm.

\subsubsection{Comment on  the variance reduction:} 


While estimating $\F_2$ note that when the original variance in AMS-sketch is large, i.e., when $(\mathbf{F}_2^2 - \mathbf{F}_4)$ is large (see Theorem~\ref{thm:AMS}), then
many of the $f_i$s are also expected to be large.\footnote{Otherwise, for instance in an extreme case, when there is only single large $f_i$ then  $(\mathbf{F}_2^2 - \mathbf{F}_4) = 0$} In this case $\F_1^2 - \F_2 = \sum_{i \neq j} f_i f_j $ would also be large. Hence we would expect bigger variance reduction in absolute terms given by ${\left(\F_1^2-\F_2\right)^2}/{\F_0(\F_0-1)}$ due to Theorem~\ref{thm:AMS}.

\begin{figure}[!ht]
\centering
 \includegraphics[height=4.5cm,width=0.8\linewidth]{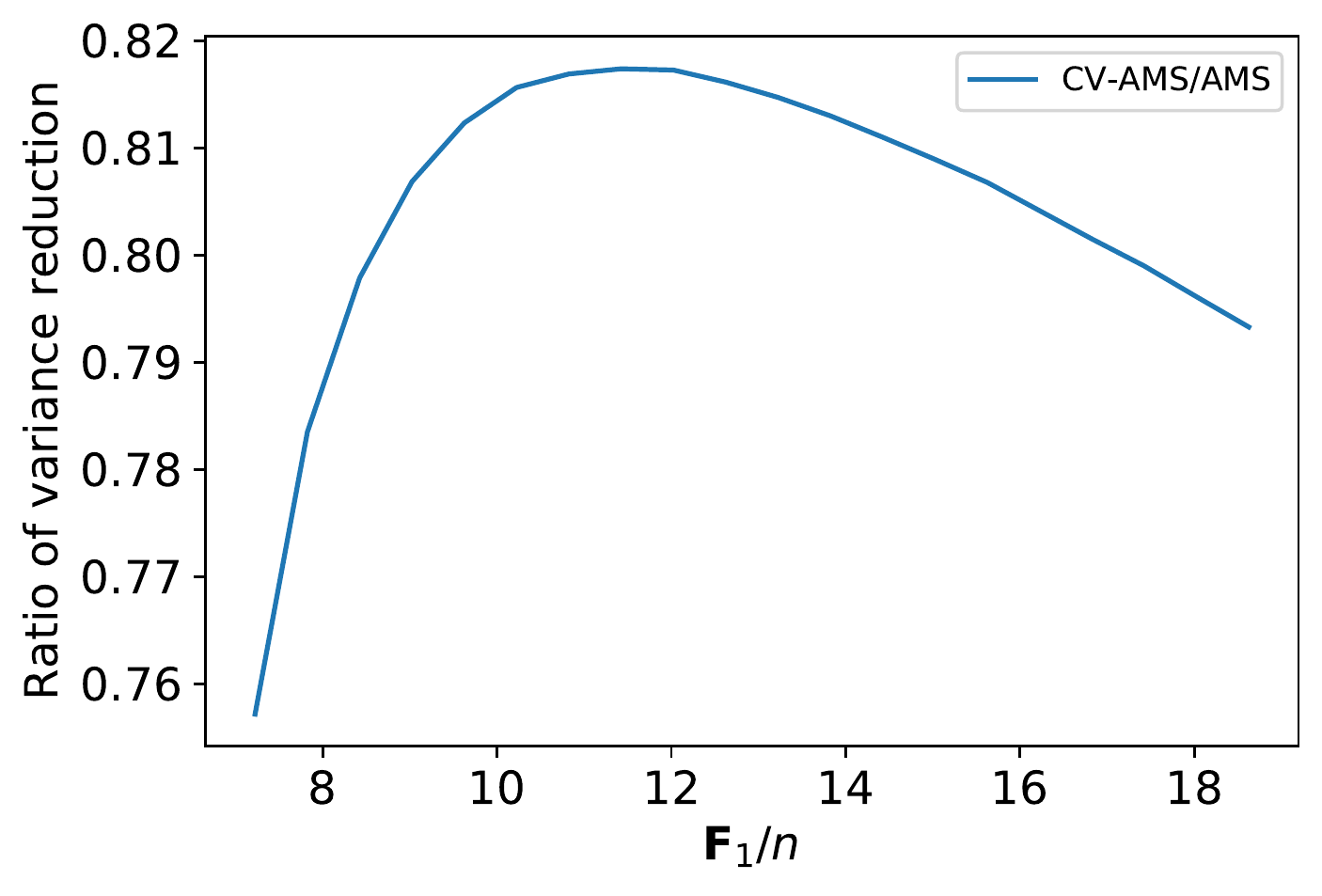}
 \vspace{-5mm}
\caption{The ratio of the variance between our proposal (CV-AMS) with that of AMS sketch in the estimation of  $\mathbf{F}_2$  of the data stream -- a smaller value indicates that our proposal has a smaller variance than that of AMS-sketch. $X$-axis corresponds to  $\mathbf{F}_1/n$.
 }
\label{fig:theoritical_variance_F2}
\end{figure}

We also wish to get a visual understanding of the variance reduction in $\mathbf{F}_2$ estimation. To do so, we generate a  random data stream having  $1000$ distinct items \textit{s.t.} each item has a random frequency between $1$ and $10$. Then, using Theorem~\ref{thm:AMS}, we calculate the variance of our proposal (CV-AMS) and AMS sketch in the estimation of  $\mathbf{F}_2$,  and compute their ratio  -- a smaller value is an indication of better performance. We plot this ratio by varying   $\mathbf{F}_1/n$ -- keeping the number of distinct items constant but increasing their respective frequencies for randomly sampled items. We summarise our observation in Figure~\ref{fig:theoritical_variance_F2}. We notice that the ratio of these two variances is small which indicates that CV-AMS has a smaller variance than that of vanilla AMS-sketch. 

Similarly, in the estimation of  the inner product of the frequency vector when the original variance in AMS-sketch is large, i.e., when $\sum_{i\neq j}{f_i}^2{g_j}^2$ is large (see Theorem~\ref{thm:AMS_IP_estimation}), then
many of the $f_i$s and $g_i$s are also expected to be large. Therefore we get bigger variance reduction in absolute terms given by  ${2\left(\langle \mathbf{f}, \mathbf{g} \rangle(\F_2+\G_2)\right)^2}/{2\langle \mathbf{f}, \mathbf{g} \rangle^2+{\F_2}^2+{\G_2}^2}$. 

\begin{figure*}[!ht]
\centering
 \includegraphics[height=3.5cm,width=\linewidth]{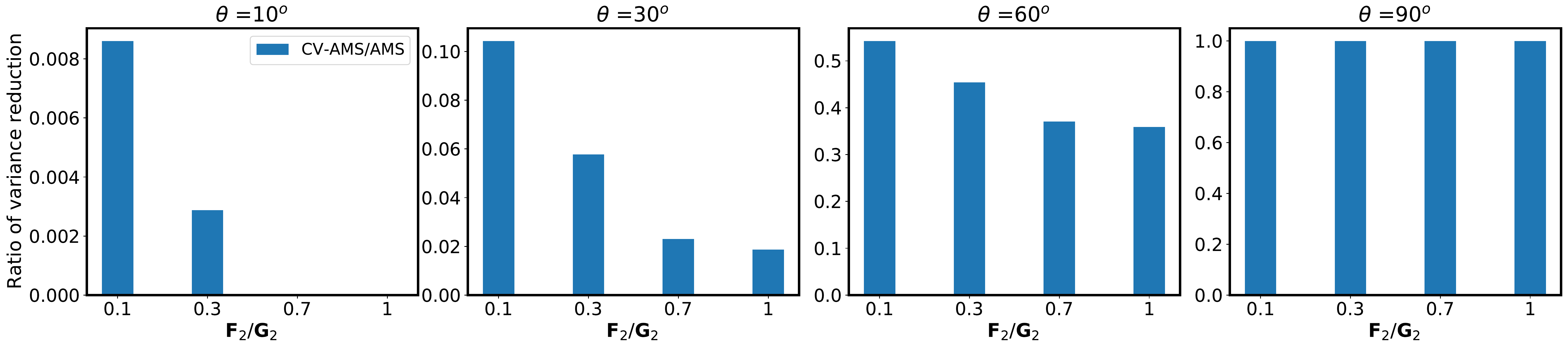}
 \vspace{-5mm}
\caption{ The ratio of the variance of our proposal (CV method) with that of AMS sketch in the inner product estimation of a pair of data streams $\sigma_1$ and $\sigma_2$ -- a lower value indicates that the CV method has a smaller variance. $\theta$ denote the angle between the corresponding frequency vectors $\mathbf{f}$ and $\mathbf{g}$ of the data streams  $\sigma_1$ and $\sigma_2$, respectively. 
}
\label{fig:theoritical_variance_IP}
\end{figure*}

We also wish to visually analyze the variance reduction in the inner product estimation of a pair of the data streams using our proposal (CV method) and AMS sketch.  To do so we generate a random pair of streams having $1000$ distinct items and the frequency of each item is randomly sampled between $1$ and $100$. We also vary the angle (denoted by $\theta$) between data stream and their respective $\ell_2$ norms (denoted by $\F_2$ and $\G_2$ of data streams $\sigma_1$ and $\sigma_2$ respectively). We choose the angle $\theta \in \{10^{o}, 30^{o}, 60^{o}, 90^{o}\}$, the ratio    $\mathbf{F}_2/\mathbf{G}_2 \in \{ 0.1,0.4,0.7,1\}$. We compute the corresponding variances of our proposal (CV method) and AMS sketch and record their ratio, and summarise it in Figure \ref{fig:theoritical_variance_IP}. Note that a smaller value of this ratio is an indication of smaller variance  by our proposal.  We notice that at smaller values of $\theta$ we obtained a much higher variance reduction because the inner product of the corresponding frequency vector is higher, whereas when $\theta=90^{o}$, our proposal doesn't provide any variance reduction. Further, we obtain a higher variance reduction when the ratio of $\F_2$ and $\G_2$ is close to $1$.  

We note that there are some results obtaining further improvement of AMS-sketch for $\F_2$~\cite{KaneNPW11, KaneNW10}, and for $\F_k$, with $k>2$~\cite{IndykW05,BhuvanagiriGKS06,Ganguly09,GangulyW18}.  However, the focus of this work is to demonstrate variance reduction for frequency estimation via control variate method. We believe that our technique can also be applied to the improvements of the AMS-sketch to obtain similar variance reductions. We state it as an open question of the work.


\noindent \textbf{Known applications of control variates:} Control variate technique has been used recently for reducing the variances of the estimates obtained in several Monte-Carlo simulations. Kang \textit{et. al.}~\cite{KangH17, kang} used it for improving the estimates for inner product and Euclidean distance obtained from random projection.  However, to the best of our knowledge this approach has not been tried for reducing the variance of the streaming algorithm. In this work, we initiate this study. 

\noindent\textbf{Organization of the paper:} The rest of the paper is organized as follows: in Section~\ref{sec:background}, we state some definitions  and known facts which are used in the paper, also for the sake of completeness of the paper we briefly revisit the algorithms of AMS-sketch for $\F_2$~\cite{AMS} and estimating inner product of frequency vectors of a pair of data streams,  and their analysis. In Section~\ref{sec:analysis}, we give proofs of our results stated in Theorem~\ref{thm:AMS} and Theorem~\ref{thm:AMS_IP_estimation}. In Section~\ref{sec:experiments}, we complement our theoretical results with experiments on synthetic and real-world datasets. Finally in Section~\ref{sec:conclusion}, we conclude our discussion and state some potential open questions of the work.

\section{Background}\label{sec:background}



\begin{table*}
\caption{{Notations}\label{tab:notations}} 
    \centering
    \scalebox{1}{
      \noindent\begin{tabular}{lc||lc}
    \hline
    Set $\{1, 2, \ldots n\}$ & $[n]$ & Data stream $\{a_1, a_2, \ldots, a_{m_1}\}$  & $\sigma_1$ \\
    \hline
    Stream elements  &$\forall i\in[m_1], j\in[m_2], a_i, b_j\in[n]$ & Data stream $\{b_1, b_2, \ldots, b_{m_2}\}$  & $\sigma_2$ \\
    \hline
    Freq. vector of $\sigma_1$  &$\mathbf{f}=\langle f_1, \ldots f_i\ldots f_n \rangle$ \textit{s.t.} $f_i:=$ freq. of $i$-th item  & Sketch of data stream $\sigma_1$ &$\tilde{f}$ \\
    \hline
    Freq. vector of $\sigma_2$  &$\mathbf{g}=\langle g_1, \ldots g_i. \ldots, g_n \rangle$ \textit{s.t.} $g_i:=$ freq. of $i$-th item  & Sketch of data stream $\sigma_2$ &$\tilde{g}$ \\
    \hline
    Hash function    &$h:[n]\mapsto \{-1, +1\}$& $h(i):i\in [n]$  &$Y_i$ \\
    \hline
     $\sum_{i=1}^n{f_i}^k$& $\F_k$& $\sum_{i=1}^k{f_i}{g_i}$  &$\langle \mathbf{f}, \mathbf{g} \rangle$ \\
    \hline
\end{tabular}
}
\end{table*}

\begin{defi}[$k$-Universal Hashing]\label{def:four_universal} 
A randomized function $h:[n]\mapsto [l]$ is $k$-universal if $\forall i_1\neq i_2\ldots,\neq i_k \in[n]$,  the following holds true   for any $z_1, z_2, \ldots z_k \in [k]$:
\[
\Pr[h(i_1)=z_1, \text{and~} h(i_2)=z_2, \text{~and~} \ldots h(i_k)=z_k]=\frac{1}{l^k}.
\]
A simple $k$-universal hash function example is the following~\cite{k-universal}:
\[
h(x)=\left(\left[\sum_{i=0}^{k-1}a_ix^i\right]\mod p \right)\mod l,  
\]
where $p$ is a large prime number, and $a_i$'s are some randomly sample positive integers smaller than $p$.
\end{defi}


\begin{lem}[The Median-of-Means Improvement~Lemma $4.4.1$ of~\cite{amit_notes}]\label{lem:median_improvement}
 There is a universal constant $c$ such that the following holds. Let random variable $X$ be an unbiased estimator of a real  quantity $Q$. Let $\{X_{ij}\}_{i\in [t], j\in [k]}$ be collection of random variables with each $X_{ij}$ distributed identically to $X$, where
 \[
 t=c\log\left(\frac{1}{\delta}\right)~\text{and~} k=\frac{3\Var[X]}{\varepsilon^2 \E[X]^2}.
 \]
 Let $Z=median_{i\in [t]}\left(\frac{1}{k} \sum_{j=1}^k X_{ij}\right)$. Then we have $\Pr[|Z-Q|\geq \varepsilon Q]\leq \delta$. That is, $Z$ is an $(\varepsilon,\delta)$ estimator for $Q$. This if an algorithm can produce $X$ with $s$ bits of space then there is an $(\varepsilon,\delta)$-estimation algorithm using 
 \[
 O\left(s\cdot \frac{\Var[X]}{\varepsilon^2 \E[X]^2}\cdot \frac{1}{\varepsilon^2}\log \frac{1}{\delta} \right).
 \]
 bit of space.
\end{lem}
\let\thefootnote\relax\footnotetext{Discrimilar: We adapt a  similar writing style as of~\cite{amit_notes} for describing Algorithm~\ref{alg:AMS}  and its  proof of correctness stated in Theorem~\ref{thm:AMS_vanilla}. 
}

We require the following results from probability theory in order to proof some of our results.

\begin{lem}~\cite{provost1992quadratic}\label{lem:murihead}
Let $\mathbf{w}\sim \mathcal{N}(\mathbf{0}, \mathbf{\Sigma})$, and $\mathbf{A, B}$ are symmetric matrices, then
\begin{align*}
    \Var[\mathbf{w^TAw}]&=2\Tr[\mathbf{A\Sigma A\Sigma}],\\
    \Cov[\mathbf{w^TAw, w^TBw}]&=2\Tr[\mathbf{A\Sigma B\Sigma}],
    \end{align*}
where $\Tr$ is the trace operator of the matrix.
\end{lem}
\begin{thm}[Multivariate Lyapunov CLT~\cite{feller1}]\label{thm:multivariate_Lyapunov_CLT}
Let  $\{\mathbf{X}_{1},\ldots \mathbf{X}_{n}\}$ be a sequence of independent random vectors such that each entry of the both \begin{inparaenum}[(i)]\item the expected value of the random vector $\{\mathbf{X}_{i}\}_{i=1}^n$, \item and the corresponding covariance matrix $\mathbf{\Sigma}_{i}$ \end{inparaenum}, is finite.  We define 
$$\mathbf{V}_{n} = \sum_{i=1}^{n} \mathbf{\Sigma}_{i}.$$
If for some $ \delta >0$  the following condition holds true
\begin{align*}
  &\lim_{n \to \infty} ||\mathbf{V}_{n}^{\frac{1}{2}}||^{2+\delta} \sum_{i=1}^{n} \E\left[||\mathbf{X}_{i} - \E[\mathbf{X}_{i}]||^{2 +\delta} \right] = 0,~\text{then}  \\
  &\V_{n}^{-\frac{1}{2}} \sum_{i=1}^{n} (\mathbf{X}_{i} -\E[\mathbf{X}_{i}]) \overset{d}{\to} \mathcal{N}(\mathbf{0}, \mathbf{I}),
\end{align*}
as $n$ tends  to infinity. Where  $\overset{d}{\to}$ denotes the convergence in distribution;  $\mathbf{0}$ and  $\mathbf{I}$  denote vector with each entry equal to zero and  the identity matrix, respectively. 
\end{thm}

\subsection{Revisiting AMS-Sketch~\cite{AMS} for $\F_2$ -- Tug-of-war Sketch.} 
Alon, Matias and Szegedy~\cite{AMS} give a sketching algorithm for computing $\F_2$ norm of frequency vector of data stream. This algorithm is also popularly known as \textit{``Tug-of-war"} sketch. We discuss it in Algorithm~\ref{alg:AMS}, and Theorem~\ref{thm:AMS_vanilla} discuss its theoretical analysis. 

\begin{algorithm}
\caption{AMS-Sketch~\cite{AMS} -- Tug-of-war Sketch}\label{alg:AMS}
\begin{algorithmic}
 \State \textbf{Initialize}:
 Choose a random  hash function $h:[n]\mapsto \{-1, +1\}$,  from a $4$-universal family
 \State $x\leftarrow 0$
\State \textbf{Process}$(j,c)$:
\State $x\leftarrow x+c \cdot h(j)$
 \State \textbf{Output:}
\State  $x$
\end{algorithmic}
\end{algorithm}



We now give an analysis on the guarantee offered by  AMS-Sketch. Let $\tilde{X}$ denote the value of $x$, when the algorithm has finished processing the stream $\sigma$. Then we have 
\[
\tilde{X}=\sum_{j\in[n]}f_jh(j).\numberthis \label{eq:rv}
\]
\begin{thm}[Adapted from the results of~\cite{AMS}]\label{thm:AMS_vanilla} 
Let $\tilde{X}$ be the random variable denoting the value of $x$ when the algorithm finishes processing the stream $\sigma$, and  let us denote  $X=\tilde{X}^2$. Then we have the following
\begin{itemize}
    \item $\E[X] =\F_2$, 
    \item $\Var[X]=2(\F_2^2-\F_4)$.
\end{itemize}
\end{thm}

\subsection{Computing inner product of two frequency vectors}
As an application of the control variate trick, we provide significant variance reduction in the frequency moments estimators of AMS-sketch.  Algorithm~\ref{alg:AMS} can also be used to compute the inner product of frequency vectors corresponding to two data streams.  Let $\sigma_1, \sigma_2$ are two data streams of lengths $m_1$ and $m_2$, respectively,  where each element of both the streams belong to the  universe  $[n]:=\{1, 2, \ldots, n\}$. 
Let   $\mathbf{f}=\langle f_1, f_2, \ldots, f_n \rangle$  and $\mathbf{g}=\langle g_1, g_2, \ldots, g_n \rangle$ denote the frequency vectors corresponding to the streams $\sigma_1$ and $\sigma_2$, respectively. 
Let us denote $\Tilde{f}$ and $\Tilde{g}$ the sketch of $\mathbf{f}$ and $\mathbf{g}$ obtained via Tug-of-War sketch (Algorithm~\ref{alg:AMS}), respectively, where
\begin{align*}
    \Tilde{f}&=\sum_{i=1}^n f_i h(i),\\
    \Tilde{g}&=\sum_{i=1}^n g_i h(i).  
 \end{align*}

\begin{thm}\label{thm:AMS_vanilla_dot_product} 
Suppose $\Tilde{f}$ and $\Tilde{g}$ are two random variables that are output of Algorithm~\ref{alg:AMS} after  processing the streams $\sigma_1$ and $\sigma_2$, and let us denote $ X^{(2)}:= \Tilde{f} \Tilde{g}$. Then we have the following
\begin{itemize}
    \item $\E[X^{(2)}] =\langle \mathbf{f}, \mathbf{g} \rangle$, 
    \item $\Var[X^{(2)}]=\sum_{i\neq j}{f_i}^2{g_j}^2+\sum_{i\neq j, i, j \in [n]} f_ig_if_jg_j$.
\end{itemize}
\end{thm}

\noindent \textbf{Concentration analysis and the space complexity for Tug-of-war Sketch:} 
While estimating  $\F_2$, in Algorithm~\ref{alg:AMS},  the sketch is  the final value outputted by the variable $x$. The absolute value of $x$ is at most $f_1+f_2+\ldots+f_n=m$. Therefore the space required for the sketch $x$ is $O(\log m)$. The space required to store the hash function is $O(\log n)$. Thus, the overall space requirement is $O(\log m+\log n).$ Further, the space requirement for  $(\varepsilon,\delta)$-estimator using Lemma~\ref{lem:median_improvement} is  $O\left(\frac{1}{\varepsilon^2} \log \frac{1}{\delta}(\log m+\log n) \right).$

Further, in case of inner product estimation of a pair of data streams $\sigma_1$ and $\sigma_2$, the space required to store their sketches obtained from Algorithm~\ref{alg:AMS} are $O(\log m_1)$ and $O(\log m_2)$, respectively. Therefore the overall space requirement is $O(\log m_1+\log m_2+\log n)=O(\log {m_1} {m_2}+\log n).$ Thus, the space requirement for  $(\varepsilon,\delta)$-estimator using Lemma~\ref{lem:median_improvement} is  $O\left(\frac{1}{\varepsilon^2} \log \frac{1}{\delta}(\log {m_1}{m_2}+\log n) \right).$

\section{Analysis}\label{sec:analysis}
The frequency moment estimation computed via AMS-sketches can be made to offer $(\varepsilon, \delta)$-guarantee by taking several independent copies of the random variable and computing the median of the means of the estimates by Lemma~\ref{lem:median_improvement}. Further, a reduction in the variance leads to a reduction in the number of independent copies of the random variables required for providing the same guarantee.  In what follows we give proofs of Theorems~ \ref{thm:AMS}, where we perform the variance reduction analysis for one estimate only.

\subsection{Improving the variance bounds  of AMS-Sketch~\cite{AMS} for $\F_2$ using control variate trick -- Proof of Theorem~\ref{thm:AMS}} 
\begin{proof}
Recall that in AMS-Sketch, we denote the hash function $h(j)$ with the  random variable $Y_j$. We now define our control variate random variable as follows:
\begin{align*}
Z&=\sum_{i\neq j, i,j\in [n]} Y_iY_{j}.\numberthis \label{eq:CV_AMS}\\
&=\left(\sum_{i\in [n]} Y_i\right)^2-\sum_{i\in [n]} Y_i^2 =\left(\sum_{i\in [n]} Y_i\right)^2-n. \numberthis \label{eq:CV_AMS1}\\
\end{align*}
In the following analysis we will repeatedly use the following equalities:   $\E[Y_j]=0$, $\E\left[ Y_iY_{j} \right]=0$ and $\E[Y_j^2]=\E[1]=1.$ We  first calculate the expected value of the random variable $Z$.
\begin{align*}
    \E[Z]&=\E\left[\sum_{i\neq j, i,j\in [n]} Y_iY_{j} \right]=\sum_{i\neq j, i,j\in [n]}\E\left[ Y_iY_{j} \right]=0.\numberthis \label{eq:exp_z_AMS}
\end{align*}

We now calculate the variance of the random variable $Z$.
\begin{align*}
    &\Var[Z]=\Var\left[\sum_{i\neq j, i,j\in [n]} Y_iY_{j}\right].\\
    &=\sum_{i\neq j, i,j\in [n]}\Var\left[ Y_iY_{j}\right]+\sum_{\begin{subarray}{l}i\neq j\neq l\neq m,\\ i,j,l,m \in [n]\end{subarray}}\Cov(Y_iY_j, Y_lY_m).\\
    &=\sum_{i\neq j, i,j\in [n]} \left(\E[Y_i^2Y_{j}^2]- \E[Y_iY_{j}]^2\right)+\\&+\sum_{i\neq j\neq l\neq m, i,j,l,m \in [n]} \left(\E[Y_i Y_jY_lY_m]-\E[Y_i Y_j]\cdot\E[Y_lY_m] \right).\\
    &=\sum_{i\neq j, i,j\in [n]} \left(\E[1]- \E[Y_iY_{j}]^2\right)+\\&+\sum_{i\neq j\neq l\neq m, i, j, l, m \in [n]} \left(\E[Y_i Y_jY_lY_m]-\E[Y_i Y_j]\cdot\E[Y_lY_m] \right).\\
     &=\sum_{i\neq j, i,j\in [n]} 1- 0+(0-0).\\
    &=n(n-1)=\F_0(\F_0-1).\numberthis \label{eq:Var_AMS}
\end{align*}
We now calculate the covariance between random variable $X$ and control variate random variable $Z$.

\begin{align*}
    &\Cov(X, Z)\\
	&=\Cov\left(\sum_{j\in[n]}{f_j}^2 +\sum_{i\neq j,i,j\in[n]}{f_i}{f_j}{Y_i}{Y_j}, \sum_{\begin{subarray}{l}l\neq m,\\l,m\in[n]\end{subarray}}{Y_l}{Y_m}\right). \numberthis\label{eq:eq200}\\
    &=\Cov\left(\sum_{i\neq j,i,j\in[n]}{f_i}{f_j}{Y_i}{Y_j}, \sum_{l\neq m,l,m\in[n]}{Y_l}{Y_m}\right).\\
    &=\E\left[\left(\sum_{i\neq j,i,j\in[n]}{f_i}{f_j}{Y_i}{Y_j}\right)\cdot\left(\sum_{l\neq m,l,m\in[n]}{Y_l}{Y_m}\right)\right]\ldots\\&~~-\E\left[\sum_{i\neq j,i,j\in[n]}{f_i}{f_j}{Y_i}{Y_j}\right]\cdot\E\left[\sum_{l\neq m,l,m\in[n]}{Y_l}{Y_m}\right].\\
    &=\E\left[\left(\sum_{i\neq j,i,j\in[n]}{f_i}{f_j}{Y_i}{Y_j}\right)\cdot\left(\sum_{l\neq m,l,m\in[n]}{Y_l}{Y_m}\right)\right].\\
    &=\E\left[\sum_{i\neq j,i,j\in[n]}{f_i}{f_j}{{Y_i}}^2{{Y_j}}^2\right]\ldots\\&+\E\left[\sum_{i\neq j\neq l\neq m,i,j,l,m\in[n]}{f_i}{f_j}{{Y_i}}{{Y_j}}{{Y_l}}{{Y_m}}\right].\\
     &=\sum_{i\neq j,i,j\in[n]}{f_i}{f_j}\E\left[{{Y_i}}^2{{Y_j}}^2\right]\ldots\\&+\sum_{i\neq j\neq l\neq m,i,j,l,m\in[n]}{f_i}{f_j}\E\left[{{Y_i}}{{Y_j}}{{Y_l}}{{Y_m}}\right].\\
        &=\sum_{i\neq j,i,j\in[n]}{f_i}{f_j}\E\left[1\right]+\sum_{i\neq j\neq l\neq m,i,j,l,m\in[n]}{f_i}{f_j}\times 0.\\
       &=\sum_{i\neq j,i,j\in[n]}{f_i}{f_j}.\\&=\left(\sum_{i\in[n]}f_i\right)^2-\sum_{i\in[n]}f_i^2= \F_1^2-\F_2.
    \numberthis \label{eq:cov_AMS}
\end{align*}


Equations~\ref{eq:cov_AMS} along with Equation~\ref{eq:Var_AMS} give the control variate coefficient $\hat{c}$ and variance reduction as follows. This concludes a proof of the theorem.
\begin{align*}
    &\hat{c}=-\frac{\Cov[X,Z]}{\Var[Z]}=-\frac{\F_1^2-\F_2}{\F_0(\F_0-1)}. \numberthis \label{eq:control_variate_CMS}\\
    &\text{Variance Reduction}=\frac{\Cov[X,Z]^2}{\Var[Z]}=\frac{\left(\F_1^2-\F_2\right)^2}{\F_0(\F_0-1)}.\numberthis \label{eq:variance_reduction_CMS}
\end{align*}

\end{proof}

\noindent\textbf{How to compute $Z$:}
Recall that (from Equation~\ref{eq:CV_AMS1}) our control variate random variable 
 $Z=\left(\sum_{i\in [n]} Y_i\right)^2-n$, where $Y_j:=h(j).$
 Note that  $Z$  only depends on the number of distinct elements, and is independent of the data stream $\sigma$. It can be easily computed (via Equation~\ref{eq:CV_AMS1}) by examining the hash function $h:[n]\mapsto \{-1, +1\}$ and maintaining a counter for the summation $\sum_{i\in [n]} Y_i$.

\begin{remark}
In this work, we use AMS sketch for estimating the $\ell_2$ norm of the frequency vector, where each $f_i$ takes a non-negative value. However, we note that AMS sketch also works for any real valued vector, say $\mathbf{u}=\langle u_1, u_2, \ldots, u_n \rangle$, where $\mathbf{u} \in \R^d$   ($u_i$ can take negative values as well). We remark that our algorithm also works in this scenario as well. However, in this case, we assume that we know the value of $\sum_{i=1}^n u_i$ and $\sum_{i=1}^n u_i^2$ to compute the value of $\hat{c}$ (see Equations~\eqref{eq:cov_AMS},\eqref{eq:control_variate_CMS}).
\end{remark}

\subsection{Variance reduction in Tug-of-War estimator for dot product using Control variate -- Proof of Theorem~\ref{thm:AMS_IP_estimation}:} 

Let $\mathbf{f}=\langle f_1,\ldots, f_n \rangle$ and $\mathbf{g}=\langle g_1,\ldots, g_n \rangle$ be two frequency vectors corresponding to two data streams $\sigma_1$ and $\sigma_2$, and $\tilde{f}, \tilde{g}$ are their sketches obtained from tug-of-war sketch algorithm (Algorithm~\ref{alg:AMS}). Recall that our random variable for  estimating the inner product between $\mathbf{f}$ and $\mathbf{g}$ is:
\begin{align}
    X^{(2)}&=\Tilde{f}\Tilde{g}.\\
    \E[X^{(2)}]&=\langle \mathbf{f}, \mathbf{g}\rangle. \label{eq:cv_dot_product}
\end{align}

We define our control variate random variable as follows:
\begin{align}
    Z^{(2)}&=\Tilde{f}^2+\Tilde{g}^2.\label{eq:cv_dot_product_defintion}\\
    \E[Z^{(2)}]&=\F_2+\G_2. \label{eq:cv_dot_product}
\end{align}
Equation~\eqref{eq:cv_dot_product} hold due to Theorem~\ref{thm:AMS_vanilla_dot_product} and linearity of expectation. 

We now compute the covariance between control variate random variables $Z^{(2)}$ and our estimator  $X^{(2)}$, and the variance of $Z^{(2)}$. We aim to compute it using Lemma~\ref{lem:murihead}. However, in order to use this lemma we need to prove that the joint distribution of $\tilde{f}$ and $\tilde{g}$ is bi-variate normal. We show this under the convergence in distribution and as $n\to \infty$. We use the multivariate lyapunov central limit theorem (stated in Theorem~\ref{thm:multivariate_Lyapunov_CLT}) for the same. We show it in the following theorem, and the subsequent corollary. We defer their proofs due to the space constraints.

\begin{thm}\label{thm:biviariate_gaussian}
If $~\forall i,  1\leq i\leq n,$  $\E\left[(f_i^2+g_i^2)^{\frac{2+\delta}{2}}\right]$ take finite nonzero value   for some $\delta>0$,  then as $n \to \infty$,   we have
\begin{equation*}
\mathbf{\V_n}^{-\frac{1}{2}}
\begin{bmatrix}
\tilde{f}  \\
\tilde{g} \\
\end{bmatrix}
\overset{d}{\to} \mathcal{N} \left( \mathbf{0},
\mathbf{I}
 \right). 
  \end{equation*}
  where $\overset{d}{\to}$ denotes convergence in distribution, 
 $\mathbf{\V_n}=
\begin{bmatrix}
                    \F_2 & \langle \mathbf{f}, \mathbf{g} \rangle\\
                   \langle \mathbf{f}, \mathbf{g} \rangle& \G_2 
                    \end{bmatrix}_{2 \times 2}$, $\mathbf{0}$ is a $(2 \times 1)$ dimensional vector with each entry as  zero,   $\mathbf{I}$  is an $(2 \times 2)$
                    identity matrix.

\end{thm}

\begin{cor}\label{cor:cor_bivariate}
Following the assumption stated in Theorem~\ref{thm:biviariate_gaussian},  
\begin{align*}
   \V_{n}^{-1/2}\begin{bmatrix}
\tilde{f} \\
\tilde{g}
\end{bmatrix}
\overset{d}{\to}  \mathcal{N} \left( \mathbf{0}, \mathbf{I}
 \right)
 & \implies
 \begin{bmatrix}
\tilde{f}  \\
\tilde{g} \\
\end{bmatrix}
\overset{d}{\to} \mathcal{N} \left( \mathbf{0},
\mathbf{V}_{n} \right). \numberthis \label{eq:normal_dist}
\end{align*}
\end{cor}

In the following, we conclude a proof of Theorem~\ref{thm:AMS_IP_estimation} using the results stated in Theorem~\ref{thm:biviariate_gaussian}, Corollary~\ref{cor:cor_bivariate} and Lemma~\ref{lem:murihead}. 

\subsubsection*{Proof of Theorem~\ref{thm:AMS_IP_estimation}:}
\begin{proof}
    Let $X^{(2)}$ be the random variable denoting the estimate of our interest, and $Z^{(2)}$ be our control variate estimate. 
\begin{align}
    X^{(2)}&:=\tilde{f}\tilde{g}= \begin{bmatrix} 
\tilde{f} & \tilde{g} \\
\end{bmatrix} \begin{bmatrix} 
0 & \frac{1}{2}  \\
\frac{1}{2} & 0 \\
\end{bmatrix} \begin{bmatrix} 
\tilde{f}\\
\tilde{g}\\
\end{bmatrix}.\\
    Z^{(2)}&:=\tilde{f}^2+\tilde{g}^2=\begin{bmatrix} 
\tilde{f} & \tilde{g} \\
\end{bmatrix} \begin{bmatrix} 
1 & 0 \\
0 & 1 \\
\end{bmatrix} \begin{bmatrix} 
\tilde{f}\\
\tilde{g}\\
\end{bmatrix}.\numberthis\label{eq:cv_rand_ip}
\end{align}
From Corollary~\ref{cor:cor_bivariate}, we have
\begin{align*}
\begin{bmatrix}
\tilde{f}  \\
\tilde{g} \\
\end{bmatrix}
\overset{d}{\to} \mathcal{N} \Bigg( \begin{bmatrix}
0 \\
0 \\
\end{bmatrix},
\Sigma=
\begin{bmatrix}
\F_2  & \langle \mathbf{f}, \mathbf{g} \rangle\\
\langle \mathbf{f}, \mathbf{g} \rangle & \G_2\\
\end{bmatrix}
 \Bigg). \numberthis \label{eq:normal_dist_ip}
\end{align*}
Using Lemma~\ref{lem:murihead}, we  calculate the co-variance between $X^{(2)}$ and $Z^{(2)}$, and the variance of $Z^{(2)}$ as follows:
\begin{align*}
    &\Cov[X^{(2)}, Z^{(2)}]\\&=2 \Tr \left[ \begin{bmatrix} 
0 & \frac{1}{2}  \\
\frac{1}{2} & 0 \\
\end{bmatrix} \begin{bmatrix}
\F_2  & \langle \mathbf{f}, \mathbf{g} \rangle\\
\langle \mathbf{f}, \mathbf{g} \rangle & \G_2\\
\end{bmatrix} \begin{bmatrix} 
1 & 0 \\
0 & 1 \\
\end{bmatrix}\begin{bmatrix}
\F_2  & \langle \mathbf{f}, \mathbf{g} \rangle\\
\langle \mathbf{f}, \mathbf{g} \rangle & \G_2\\
\end{bmatrix}\right]\\
&=2\langle \mathbf{f}, \mathbf{g} \rangle(\F_2+\G_2).\numberthis\label{eq:cov_dot_prodcut_new}
\end{align*}
\begin{align*}
    &\Var[Z^{(2)}]\\&=2 \Tr \left[ \begin{bmatrix} 
1 & 0  \\
0 & 1 \\
\end{bmatrix} \begin{bmatrix}
\F_2  & \langle \mathbf{f}, \mathbf{g} \rangle\\
\langle \mathbf{f}, \mathbf{g} \rangle & \G_2\\
\end{bmatrix} \begin{bmatrix} 
1 & 0 \\
0 & 1 \\
\end{bmatrix}\begin{bmatrix}
\F_2  & \langle \mathbf{f}, \mathbf{g} \rangle\\
\langle \mathbf{f}, \mathbf{g} \rangle & \G_2\\
\end{bmatrix}\right]\\
&= 2(2\langle \mathbf{f}, \mathbf{g} \rangle^2+{\F_2}^2+{\G_2}^2).\numberthis\label{eq:var_dot_prodcut_new}
\end{align*}

Equations~\eqref{eq:cov_dot_prodcut_new} and \eqref{eq:var_dot_prodcut_new} gives us control variate coefficient and variance reduction as follows:
\begin{align*}
    \hat{c}&=-\frac{\Cov[X^{(2)},Z^{(2)}]}{\Var[{Z^{(2)}}]}=-\frac{\langle \mathbf{f}, \mathbf{g} \rangle(\F_2+\G_2)}{(2\langle \mathbf{f}, \mathbf{g} \rangle^2+{\F_2}^2+{\G_2}^2)}.\\
    &\text{Variance reduction}=\frac{\Cov[X^{(2)},Z^{(2)}]^2}{\Var[{Z^{(2)}}]}\\&=\frac{2\left(\langle \mathbf{f}, \mathbf{g} \rangle(\F_2+\G_2)\right)^2}{2\langle \mathbf{f}, \mathbf{g} \rangle^2+{\F_2}^2+{\G_2}^2}.
\end{align*}
\end{proof}

\noindent\textbf{How to compute $Z^{(2)}$:}
Recall that (from Equation~\ref{eq:cv_dot_product_defintion}) our control variate random variable 
$Z^{(2)}=\Tilde{f}^2+\Tilde{g}^2.$ We can compute its value by computing the sketches of data streams $\sigma_1$ and $\sigma_2$ (using the Tug-of-war sketch,  Algorithm~\ref{alg:AMS}), computing their squares and adding them up.

\noindent\textbf{How to choose a good control-variate function:} To obtain higher variance reduction, one should choose a control variate random variable such that it shares high covariance with a random variable corresponding to estimate, and simultaneously has low variance. Also in order to be practically useful the control variate, its expected value, variance, covariance with the original random variable should be easily computable from the dataset. We don’t claim that the variance reduction obtained by the control variate used in this paper is the best possible. It might be possible to choose a better control variate random variable and achieve higher variance reduction. We leave this as an interesting open question of the work.

\section{From theory to practice -- experiments
}\label{sec:experiments}

 We use the following  datasets for our experiments: \begin{itemize} 
 \item  \texttt{Synthetic dataset}: we generate a stream of $100000$ distinct items such that frequency of each item is randomly sampled between $1$ and $5000$. 

\item \texttt{Bag-of-word (BoW) dataset~\cite{UCI}:} This dataset consists of a corpus of documents. The raw documents were preprocessed by tokenization and removal of stopwords, then a vocabulary of unique words was generated (by keeping only those words that occurred more than ten times).  A document is represented by the frequency vector of its words, i.e., for each word in vocabulary we count the number of its occurrences in the document. For our purpose, we consider each word as an item in the stream and consider its number of occurrences in the entire corpus as its frequency. We considered the KOS dataset for our experiments which has $6906$ distinct words and $8472$ words in total. Link of the dataset is available here $^*$.
\footnote{$^{*}$ \url{https://archive.ics.uci.edu/ml/datasets/Bag+of+Words}}

\item \texttt{Transaction datasets~\cite{fimi}:} These are well-known transaction datasets discussed  in~\cite{fimi}. We considered two datasets from ~\cite{fimi} -- \texttt{T10I4D100K} and \texttt{T40I10D100K}. The former has $870$ distinct items and $1010228$ items in total, whereas the latter has $942$ distinct items, and $3960507$ items in total. Link of the datasets is available here $^{**}$.~\footnote{$^{**}$\url{http://fimi.uantwerpen.be/data/}}
\end{itemize}

\begin{figure*}[!ht]
\centering
 \includegraphics[height=3.38cm,width=\linewidth]{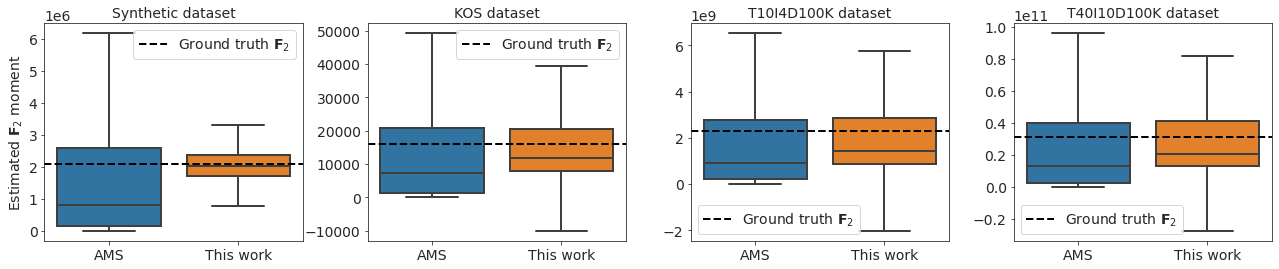}
\vspace{-5mm}
\caption{ 
Comparison between  AMS-sketch and our estimate on the variance analysis \textit{via} box-plot. 
}
\label{fig:box_plot}
\end{figure*}

\begin{figure*}[!ht]
\centering
 \includegraphics[height=3.38cm,width=\linewidth]{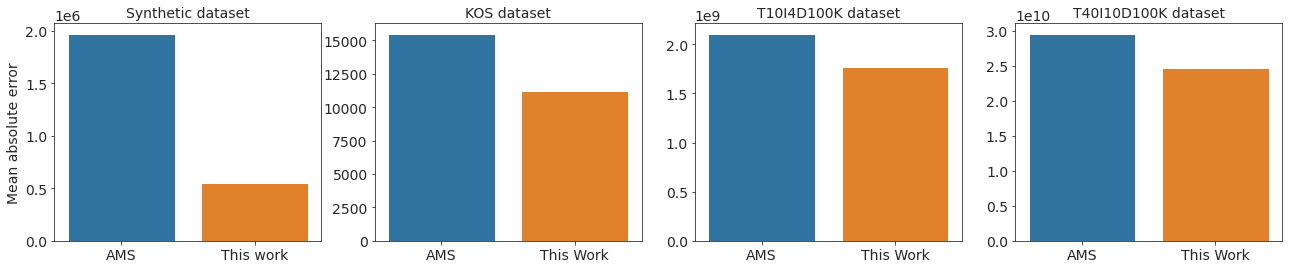}
 \vspace{-5mm}
\caption{ 
Comparison between  AMS-sketch and our estimate on the mean-absolute-error. A smaller value of mean-absolute-error is an indication of better performance. 
}
\label{fig:mean_abs_error}
\end{figure*}

\begin{figure*}[!ht]
\centering
 \includegraphics[height=3.38cm,width=\linewidth]{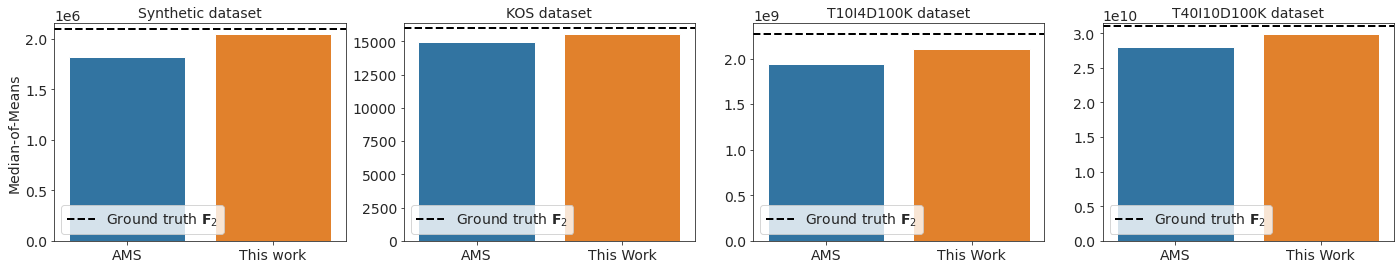}
 \vspace{-5mm}
\caption{ 
Comparison between  AMS-sketch and our algorithm on the median-of-means estimate. The dotted line corresponds to the ground truth $\F_2$. The height of the bar chart closer to the dotted line is an indication of better performance.
 }
\label{fig:median_of_mean}
\end{figure*}

\subsection{For estimating $\F_2$ using Tug-of-war sketch:}\label{subsec:F_2_experiments}
\subsubsection*{Methodology:}
 Let $X$ be the random variable denoting the estimate of $\F_2$ obtained from the AMS-sketch. Then the updated estimate proposed by our algorithm is $X+{c}(Z-\E[Z])$, where ${c}$ is the control variate coefficient, and $Z$ is the random variable denoting the control variate. 
The optimum value of $c$, for our proposal, is  given by $\hat{c}$   turns out to be 
\[
\hat{c}=-\frac{\Cov[X,Z]}{\Var[Z]}=-\frac{\F_1^2-\F_2}{\F_0(\F_0-1)}\qquad\text{(see  Equation~\ref{eq:control_variate_CMS})}.
\]

We  assume that we know the values of $\F_1$  -- the length of the stream and $\F
_0$ -- the number of distinct elements in the stream. We do not know the exact value  of $\F_2$. However we know its estimated value obtained form the AMS-sketch, which we use as a proxy.  Thus, we can compute an estimate of $\hat{c}$, which turns out to give good variance reduction as demonstrated by our experiments. 
Recall that our control variate random variable  is 
$$Z=\left(\sum_{i\in [n]} Y_i\right)^2-n \qquad\text{(see Equation~\ref{eq:CV_AMS1})}.$$ 

The  value  of  $Z$ can be easily computed by summing the hash value of each distinct item in the stream, and subtracting it with the number of distinct element in the stream.  Further, the expected value of $Z$  is  $0$ from Equation~\ref{eq:exp_z_AMS}. Therefore, we can compute the value of  $
  \text{our estimate}=  X+\hat{c}(Z-\E[Z]).
$ using the estimates of $X$, $\hat{c}$ and $Z$. We generate the $4$-universal hash functions used in the AMS algorithm following the approach stated in Definition~\ref{def:four_universal}.

 \begin{figure*}[ht]
\centering
 \includegraphics[height=3.38cm,width=\linewidth]{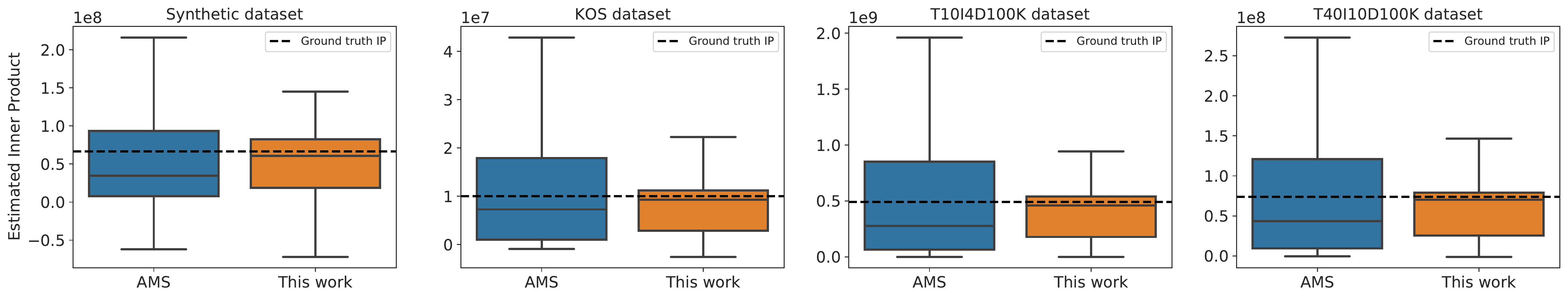}
\vspace{-5mm}
\caption{ 
Comparison between  AMS-sketch and our estimate on the variance analysis on the task of inner product  estimation \textit{via} box-plot. 
}
\label{fig:box_plot_IP}
\end{figure*}

\begin{figure*}[!ht]
\centering
 \includegraphics[height=3.38cm,width=\linewidth]{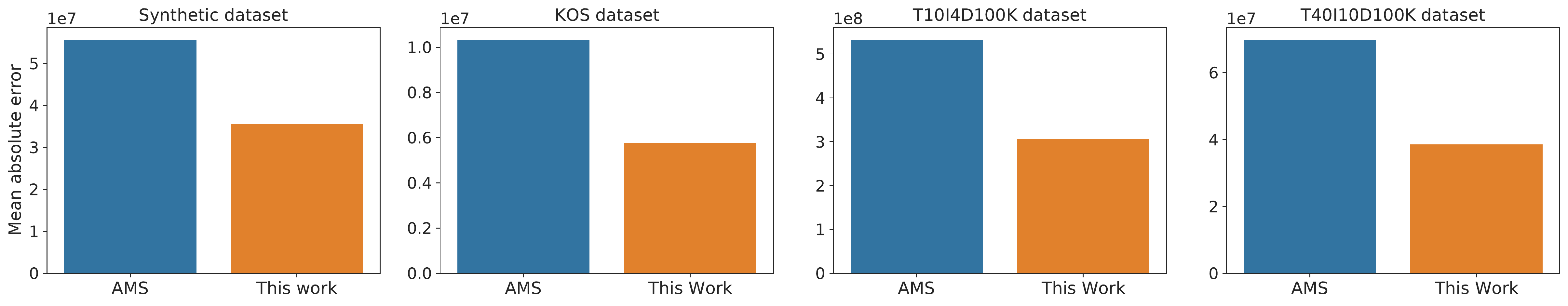}
 \vspace{-5mm}
\caption{ 
Comparison between  AMS-sketch and our estimate on the mean-absolute-error on the task of inner product estimation. A smaller value of mean-absolute-error is an indication of better performance. 
}
\label{fig:mean_abs_error_IP}
\end{figure*}

\begin{figure*}[!ht]
\centering
 \includegraphics[height=3.5cm,width=\linewidth]{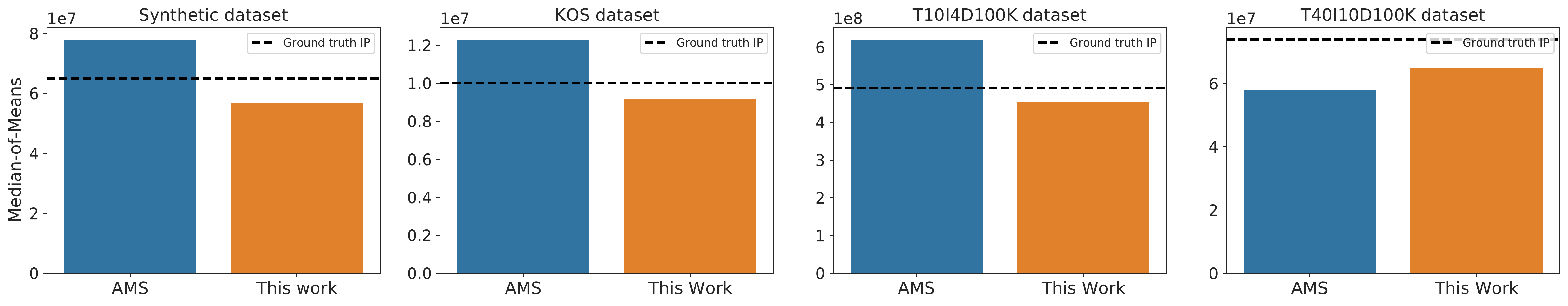}
 \vspace{-5mm}
\caption{ 
Comparison between  AMS-sketch and our algorithm on the median-of-means estimate. The dotted line corresponds to the ground truth inner product. The height of the bar chart closer to the dotted line is an indication of better performance.
 }
\label{fig:median_of_mean_IP}
\end{figure*}

\subsubsection*{Evaluation Metric:}
We evaluate the performance of our approach with the AMS-Sketch algorithm on the following three measures -- \begin{inparaenum} [(i)] \item variance analysis \textit{via} box-plot, \item mean absolute error, \item medians-of-means estimation (see Lemma~\ref{lem:median_improvement}).
\end{inparaenum}
We discuss our experimental procedure as follows:
For each dataset, we run both AMS-sketch and our proposal $1000$ times.  This gives us $1000$ different estimates for $\F_2$ both from AMS-sketch and from our method. We use these estimates to generate box-plots for variance analysis. To calculate the mean-absolute-error, we compute the absolute difference of each of these estimates with the ground truth $\F_2$, and then compute the mean of these $1000$ absolute values. A smaller value of the mean-absolute-error is an indication of better performance. In order to compute the medians-of-means estimation, we randomly make $20$ groups each with  $50$ estimates. We then compute the mean of each group and consider the median of all the $20$ means. This gives us a median-of-means estimate.  We summarise our results in Figures~\ref{fig:box_plot}, \ref{fig:mean_abs_error}, and \ref{fig:median_of_mean}, respectively.

\subsubsection*{Insight:}
In Figure~\ref{fig:box_plot}, we observe the interquartile range of our proposal is smaller than that of  AMS-sketch which implies that the variance of our proposal is smaller.  In Figure~\ref{fig:mean_abs_error}, we notice that the mean absolute error of our proposal is always smaller than that of AMS-sketch. This indicates that the error occurred in our estimate is small.
Finally, in~\ref{fig:median_of_mean}, the median-of-mean estimate of our method tends to be closer to the ground truth  $\F_2$. All these observations indicate that the variance of our estimate as well as the error in approximating the ground truth $\F_2$ are smaller than that of AMS-sketch.

\subsection{For estimating inner product between a pair of data streams  using Tug-of-war sketch:}

 \subsubsection*{Methodology:} Let $\tilde{f}$ and $\tilde{g}$ denote the sketches of a pair data streams $\sigma_1$ and $\sigma_2$ obtained via AMS-sketch. Let $X^{(2)}$ be the random variable denoting the estimate of the inner product of the corresponding frequency vectors of $\sigma_1$ and $\sigma_2$, respectively. Our control variate estimator is $X^{(2)}+\hat{c}(Z^{(2)}-\E[X^{(2)}]$, where 
 \begin{align*}
   Z^{(2)}&=\tilde{f}^2+\tilde{g}^2, \qquad (\text{see Equation}~\eqref{eq:cv_rand_ip}), \text{and}\\
    \hat{c}&=-\frac{\langle \mathbf{f}, \mathbf{g} \rangle(\F_2+\G_2)}{(2\langle \mathbf{f}, \mathbf{g} \rangle^2+{\F_2}^2+{\G_2}^2)}.
  \end{align*}
We compute the value of $Z^{(2)}$ by computing the sum of squares of the sketches of $\sigma_1$ and  $\sigma_2$ obtained via Algorithm~\ref{alg:AMS}. To compute the value of $\hat{c}$, we assume that we know the values of $\F_2$ and $\G_2$ in advance. However, we don’t know the value of  $\langle \mathbf{f}, \mathbf{g} \rangle$ -- the very quantity which we want to estimate.  For our experiments, we take the estimate obtained via AMS sketch as its proxy.

 \subsubsection*{Evaluation Metric:} We require a pair of streams to perform our experiments. We generate it as follows: for synthetic datasets, we generate a pair of streams using the similar procedure mentioned above. For BOW datasets~\cite{UCI}, recall that it is a corpus of a set of documents. We split the corpus into two equal halves consisting of the same number of documents, and we consider each half as a separate data stream. For transaction datasets, we split the streams in two equal halves and consider each half as a separate data stream.

 We compute the estimate of the inner product of a pair of input data streams $1000$ times using both AMS-sketch and our CV method. We use the same three  metrics --- \begin{inparaenum} [(i)] \item variance analysis \textit{via} box-plot, \item mean absolute error, \item medians-of-means estimation \end{inparaenum}, and summarise the corresponding plots in Figures~\ref{fig:box_plot_IP}, \ref{fig:mean_abs_error_IP}, and \ref{fig:median_of_mean_IP}, respectively.


 \subsubsection*{Insight:} Here again in Figure~\ref{fig:box_plot_IP}, we notice that our CV method has smaller variance than that of AMS-sketch.   In Figure~\ref{fig:mean_abs_error_IP}, we observe  that the mean absolute error of our proposal is always smaller than that of AMS-sketch which implies that 
our proposal has smaller errors in the inner product estimation.  Finally, in Figure~\ref{fig:median_of_mean_IP}, the median-of-mean estimate of our method tends to be closer to the ground truth  inner product. These results indicate that on the task of inner product estimation our CV proposal has smaller variance than that of AMS sketch which leads to a more accurate inner product estimation.

\section{Conclusion}\label{sec:conclusion}
 
In this work, we consider the problem of estimating the frequency moments of a large data stream, and the problem of estimating inner product between a pair of data streams . The breakthrough result due to Alon, Matias, Szegedy~\cite{AMS} gives a sublinear space algorithm for these problems. However, the variances of their estimators tend to be large when frequencies of items are large. We address this challenge and suggest a method for variance reduction at the expense of a small computational overhead. Our proposal relies on the classical control-variates~\cite{Lavenberg} method which is typically used for variance reduction in Monte-Carlo simulations. 
 
 Our work leaves several open questions and research directions: \begin{inparaenum}[\itshape a)\upshape]
\item extending our result from $\F_2$ to $\F_k$, for $k> 2$, and for $\F_0$,
\item improving the variance reduction by choosing better control variate,
\item variance reduction for other streaming algorithms.
 \end{inparaenum}
To conclude  we note that our method is simple and effective. Hence we hope that our method can be adopted in practice. 
Moreover, we believe that the control-variate trick can benefit large class streaming algorithms~\cite{amit_notes}, and randomized algorithms~\cite{motwani_raghvan_book} in general. Illustrating variance reduction for such algorithms using the control variate trick would be interesting future work.

 \bibliographystyle{plainurl}
\bibliography{MFCS2020/reference}
\onecolumn
\section{Appendix}

\subsection{Missing Proofs from Section~\ref{sec:background}:}
\subsubsection*{Proof of Theorem~\ref{thm:AMS_vanilla}:}
\begin{proof}
For brevity of the notation, we denote $Y_j=h(j)$ for each $j\in [n]$ in Equation~\ref{eq:rv}. Then we have 

$
\tilde{X}=\sum_{j\in[n]}f_jY_{j}.
$
Furthermore, by the definition of the hash function, we have $Y_{j}^2=1$,   $\E[Y_j]=0$, and $\E[Y_iY_j]=0$ for each $i\neq j, i,j\in [n]$.
\begin{align*}
  X&=\tilde{X}^2= \left(\sum_{j\in[n]}f_jY_{j}\right)^2  = \sum_{j\in[n]}f_j^2Y_{j}^2 +\sum_{i\neq j,i,j\in[n]}f_if_jY_{i}Y_{j}.\\
  X &= \sum_{j\in[n]}f_j^2 +\sum_{i\neq j,i,j\in[n]}f_if_jY_{i}Y_{j}.\numberthis \label{eq:eq100} \\
  \E[ X] &= \E\left[\sum_{j\in[n]}f_j^2 +\sum_{i\neq j,i,j\in[n]}f_if_jY_{i}Y_{j}\right].\\
  \E[ X]  &=\sum_{j\in[n]}f_j^2 +\sum_{i\neq j,i,j\in[n]}f_if_j\E\left[Y_{i}Y_{j}\right]\\
  &=\sum_{j\in[n]}f_j^2+0=\F_2. \numberthis \label{eq:AMS_exp}   
\end{align*}
 
In order to calculate the variance of $X$, we first calculate the following:
\begin{align*}
    \E[X^2]&=\E\left[\sum_{i,j,l,m \in[n]}f_if_jf_lf_m Y_iY_jY_lY_m \right].\\
    &=\E\left[\sum_{i\in[n]}f_i^4 Y_i^4 \right]+\E\left[\sum_{i\neq j, i,j\in[n]}f_i^2f_j^2 Y_i^2 Y_j^2  \right]+\E\left[\sum_{i\notin \{j,l,m\} }f_if_jf_lf_m Y_iY_jY_lY_m  \right].\\
     &=\E\left[\sum_{i\in[n]}f_i^4  \right]+\E\left[\sum_{i\neq j, i,j\in[n]}f_i^2f_j^2    \right]+\E\left[\sum_{i\notin \{j,l,m\} }f_if_jf_lf_m Y_iY_jY_lY_m  \right].\\
     &= \sum_{i\in[n]}f_i^4 + \sum_{i\neq j, i,j\in[n]}f_i^2f_j^2.\\
     &= \sum_{i\in[n]}f_i^4 + 3\left((\sum_{i\in[n]}f_i^2)^2- \sum_{i\in[n]}f_i^4\right).\\
     &=\F_4+3(\F_2^2-\F_4).
     \numberthis \label{eq:AMS_var1}   
\end{align*}
 $\E\left[\sum_{i\notin \{j,l,m\} }Y_iY_jY_lY_m  \right]=0$ because $Y_i$s are four-wise independent. 
We compute the variance of $X^2$ due to Equations~\eqref{eq:AMS_exp} and~\eqref{eq:AMS_var1}.
\begin{align*}
    \Var[X]&=\E[X^2]-\E[X]^2.\\&=\F_4 +  3(\F_2^2-\F_4)  - \F_2^2.\\&=2(\F_2^2-\F_4). \numberthis \label{eq:varinace_AMS_vannila}
\end{align*}

Equations~\eqref{eq:AMS_exp} and \eqref{eq:varinace_AMS_vannila} complete a proof of the theorem. 
\end{proof}

\subsubsection*{Proof of Theorem~\ref{thm:AMS_vanilla_dot_product}:}
\begin{proof}
Let   $\mathbf{f}=\langle f_1, f_2, \ldots, f_n \rangle$ and $\mathbf{g}=\langle g_1, g_2, \ldots, g_n \rangle$ denote two $n$-dimensional data points. We denote $\Tilde{f}$ and $\Tilde{g}$ the sketch of $\mathbf{f}$ and $\mathbf{g}$ obtained via Tug-of-War sketch, respectively. 
  \begin{align}
    \Tilde{f}&=\sum_{i=1}^n {f_i}h(i),\\
    \Tilde{g}&=\sum_{i=1}^n {g_i} h(i). 
 \end{align}
where $h(i):[n]\mapsto\{-1, +1\}$,  from a $4$-universal family. For brevity we denote $h(i)$ by $Y_i$ in the following analysis.  The estimator of Tug-of-War sketch is given as follows:
\begin{align}
   X^{(2)}:= \Tilde{f} \Tilde{g}&=\left(\sum_{i=1}^n f_i Y_i \right)\cdot \left(\sum_{i=1}^n g_i Y_i \right).\label{eq:expectation_dot_product1}\\
    &=\sum_{i=1}^n f_ig_i Y_i^2+\sum_{i\neq j, i, j \in [n]} f_ig_j Y_i Y_j.\\
    &=\sum_{i=1}^n f_ig_i+\sum_{i\neq j, i, j \in [n]} f_ig_j Y_i Y_j.\\
    \E[ X^{(2)}]&=\E[\sum_{i=1}^n f_ig_i]+\E[\sum_{i\neq j, i, j \in [n]} f_ig_j Y_i Y_j].\\
    \E[X^{(2)}]&=\sum_{i=1}^n f_ig_i+\sum_{i\neq j, i, j \in [n]} f_ig_j\E[ Y_i Y_j].\\
    \E[X^{(2)}]&=\sum_{i=1}^n f_ig_i=\langle \mathbf{f}, \mathbf{g} \rangle.\label{eq:expectation_dot_product2}
\end{align}

 We compute the variance of our estimator as follows:
\begin{align}
    \left(X^{(2)} \right)^2&=\left(\sum_{i=1}^n f_i Y_i \right)^2\cdot \left(\sum_{i=1}^n g_i Y_i \right)^2.\\
    &=\left(\sum_{i=1}^n {f_i}^2 {Y_i}^2 + \sum_{i\neq j, i, j \in [n]} f_if_j Y_i Y_j \right)\cdot \left(\sum_{i=1}^n {g_i}^2 {Y_i}^2 + \sum_{i\neq j, i, j \in [n]} g_ig_j Y_i Y_j  \right).\\
    &=\left(\sum_{i=1}^n {f_i}^2 + \sum_{i\neq j, i, j \in [n]} f_if_j Y_i Y_j \right)\cdot \left(\sum_{i=1}^n {g_i}^2  + \sum_{i\neq j, i, j \in [n]} g_ig_j Y_i Y_j  \right).\\
     &=\left(\F_2 + \sum_{i\neq j, i, j \in [n]} f_if_j Y_i Y_j \right)\cdot \left(\G_2  + \sum_{i\neq j, i, j \in [n]} g_ig_j Y_i Y_j  \right).\\
      &=\F_2\G_2 + \G_2\sum_{i\neq j, i, j \in [n]} f_if_j Y_i Y_j+ \F_2   \sum_{i\neq j, i, j \in [n]} g_ig_j Y_i Y_j+
    \sum_{i\neq j, i, j \in [n]} g_ig_j Y_i Y_j  \sum_{i\neq j, i, j \in [n]} f_if_j Y_i Y_j.\\
\E\left[\left(X^{(2)} \right)^2 \right]&=   \F_2\G_2+\E \left[\sum_{i\neq j, i, j \in [n]} g_ig_j Y_i Y_j  \sum_{i\neq j, i, j \in [n]} f_if_j Y_i Y_j \right]\\
\E\left[\left(X^{(2)} \right)^2 \right]&=   \F_2\G_2+\E \left[\sum_{i\neq j, i, j \in [n]} f_ig_if_jg_j {Y_i}^2 {Y_j}^2+  \sum_{(i,j,l,m) \in S} f_if_j g_l g_m Y_i Y_j Y_l Y_m\right],\\
\text{where } S &= \{(i,j,l,m) | (i,j) \neq (l,m), i\neq j, l \neq m, i,j,l,m \in[n]. \nonumber\\
\E\left[\left(X^{(2)} \right)^2 \right]&=  \F_2\G_2+\E \left[\sum_{i\neq j, i, j \in [n]} f_ig_if_jg_j {Y_i}^2 {Y_j}^2\right]+ \E \left[ \sum_{(i,j,l,m) \in S_1} f_if_j g_l g_m Y_i Y_j Y_l Y_m + \sum_{(i,j,l,m) \in S_2} f_if_j g_l g_m Y_i Y_j Y_l Y_m\right],\\
\text{where } S_1 &= \{(i,j,l,m)| (i,j)\neq(l,m), i=m, j=l, i \neq k , l \neq m, i,j,l,m \in [n] \} \text{ and } S_2 = S-S_1. \nonumber\\
\E\left[\left(X^{(2)} \right)^2 \right]&=  \F_2\G_2+\sum_{i\neq j, i, j \in [n]} f_ig_if_jg_j+ \E \left[ \sum_{i\neq j, i, j \in [n]} f_ig_if_jg_j {Y_i}^2 {Y_j}^2 + \sum_{(i,j,l,m) \in S2} f_if_j g_l g_m Y_i Y_j Y_l Y_m\right].\\
\E\left[\left(X^{(2)} \right)^2\right]&=   \F_2\G_2+\sum_{i\neq j, i, j \in [n]} f_ig_if_jg_j + \sum_{i\neq j, i, j \in [n]} f_ig_if_jg_j + 0.\\
\E\left[\left(X^{(2)} \right)^2\right]&=   \F_2\G_2+2\sum_{i\neq j, i, j \in [n]} f_ig_if_jg_j.\\
\end{align}

Therefore the variance of our estimator is 
\begin{align}
    \Var[\Tilde{f} \Tilde{g}]&=\E\left[\left(X^{(2)} \right)^2 \right]- \E\left[X^{(2)} \right]^2.\\
    &= \F_2\G_2+2\sum_{i\neq j, i, j \in [n]} f_ig_if_jg_j - \left(\sum_{i=1}^n f_i g_i\right)^2.\\
    &=\sum_{i\neq j}{f_i}^2{g_j}^2+\sum_{i\neq j, i, j \in [n]} f_ig_if_jg_j.\numberthis \label{eq:var_ams_dot_product}
\end{align}
Equations~\eqref{eq:expectation_dot_product2} and~\eqref{eq:var_ams_dot_product} complete a proof of the theorem. 
\end{proof}

\subsection{Missing proofs from Section~\ref{sec:analysis}:}
\subsubsection*{Proof of Theorem~\ref{thm:biviariate_gaussian}:}
\begin{proof} 
Let us denote  $\boldsymbol{\hat{\alpha}}_{i} = \left[f_{i}Y_{i}\right]^{T}$ and $\boldsymbol{\hat{\beta}}_{i} = \left[g_{i}Y_{i}\right]^{T}$. 
We define a sequence of $2$ dimensional random vectors $\{\X_i \}_{i=1}^n$, which is obtained via concatenation of vectors $\boldsymbol{\hat{\alpha}}_{i}$ and $\boldsymbol{\hat{\beta}}_{i}$
 as follows:
\begin{align*}
\X_i &= \begin{bmatrix}
\boldsymbol{\hat{\alpha}}_{i}\\
\boldsymbol{\hat{\beta}}_{i}
\end{bmatrix}=\begin{bmatrix}
{f}_i{Y}_i\\
{g}_i{Y}_i
\end{bmatrix}.\numberthis\label{eq:eq500}
\end{align*}
 
We compute the expected value and covariance matrix of the vector $\X_i$ as follows.
\begin{align*}
 \E[\X_i] &= \E 
\begin{bmatrix}
{f}_i{Y}_i\\
{g}_i{Y}_i
\end{bmatrix}  
= \begin{bmatrix}
0\\
0
\end{bmatrix}. \numberthis\label{eq:eq505}\\ &~\\
\mathbf{\Sigma}_{i} &= \Cov[\X_i]
                    =\begin{bmatrix}
                    \Cov[\boldsymbol{\hat{\alpha}}_{i}] & \Cov[\boldsymbol{\hat{\alpha}}_{i},\boldsymbol{\hat{\beta}}_{i}]\\
                   \Cov[\boldsymbol{\hat{\alpha}}_{i},\boldsymbol{\hat{\beta}}_{i}]& \Cov[\boldsymbol{\hat{\beta}}_{i}] 
                    \end{bmatrix}_{2 \times 2}\numberthis\label{eq:eq15}\\
                    &=\begin{bmatrix}
                    {f_i}^2 & {f_i}{g_i}\\
                   {f_i}{g_i}& {g_i}^2
                    \end{bmatrix}_{2 \times 2}\numberthis\label{eq:eq15122}
 \end{align*}
\noindent We  need to calculate
\begin{align*}
    \E[||\X_i - \E[\X_i]||^{2+\delta}] &= \E[||\X_i||^{2+\delta}]. \numberthis\label{eq:eqn20}
\end{align*}
To do so, we first calculate the following using  using Equation~\eqref{eq:eq500}
\begin{align*}
||\X_i||^{2} &= f_{i}^2 Y_{i}^2 + g_{i}^2 Y_{i}^2.\\ 
             &= f_{i}^2 + g_{i}^2.\\
\implies||\X_i||  &= (f_{i}^2 + g_{i}^2)^{\frac{1}{2}} .\numberthis\label{eq:eqn21}
\end{align*}
Equations~\eqref{eq:eqn20},  \eqref{eq:eqn21}, and \eqref{eq:eq505} give us the following:
\begin{align*}
    &\E\left[||\X_i - \E[\X_i]||^{2+\delta}\right] = \E\left[||\X_i||^{2+\delta}\right].\\
    &= \E \left[(f_{i}^2 + g_{i}^2)^{\frac{2 + \delta}{2}} \right]
    =  (f_{i}^2 + g_{i}^2)^{\frac{2 + \delta}{2}}. \numberthis\label{eq:eqn22}
\end{align*}
 
We now compute $\V_{n}$ using Equation~\eqref{eq:eq15}
\begin{align*}
&\V_{n} = \sum_{i=1}^{n} \mathbf{\Sigma}_{i}= \sum_{i=1}^{n} \Cov(X_{i}).\\
&= \begin{bmatrix}
\sum_{i=1}^{n} \Cov[\boldsymbol{\hat{\alpha}}_{i}] & \sum_{i=1}^{n} \Cov[\boldsymbol{\hat{\alpha}}_{i}, \boldsymbol{\hat{\beta}}_{i}]\\
\sum_{i=1}^{n} \Cov[\boldsymbol{\hat{\alpha}}_{i}, \boldsymbol{\hat{\beta}}_{i}] & \sum_{i=1}^{n} \Cov[\boldsymbol{\hat{\beta}}_{i}] 
\end{bmatrix}_{2 \times 2}.\\
&=\begin{bmatrix}
                    \sum_{i=1}^n{f_i}^2 &\sum_{i=1}^n {f_i}{g_i}\\
                   \sum_{i=1}^n{f_i}{g_i}& \sum_{i=1}^n{g_i}^2
                    \end{bmatrix}.\\
&=\begin{bmatrix}
                    \F_2 & \langle \mathbf{f}, \mathbf{g} \rangle\\
                   \langle \mathbf{f}, \mathbf{g} \rangle& \G_2 
                    \end{bmatrix}_{2 \times 2}.
  \numberthis\label{eq:eqn24}
\end{align*}

 Note that the matrix $\V_{n}$ is symmetric positive definite matrix. Hence, $\V_{n}^{-1}$ is also symmetric positive definite, and is
\begin{align}
\V_{n}^{-1} &= \frac{1}{(\F_2 \G_2  - \langle \mathbf{f}, \mathbf{g} \rangle^2)}\begin{bmatrix}
    \G_2 & - \langle \mathbf{f}, \mathbf{g} \rangle \\
    - \langle \mathbf{f}, \mathbf{g} \rangle & \F_2 \\
	\end{bmatrix}
\end{align}
We know the facts that for any matrix $\mathbf{M}$, $||\mathbf{M}||_{F}^2 = \Tr(\mathbf{M} \mathbf{M}^T) = \Tr(\mathbf{M}^T \mathbf{M})$ and for any positive definite matrix $\mathbf{P}$, there exists a unique symmetric matrix  $\mathbf{Q}$ such that $\mathbf{P} = \mathbf{Q} \mathbf{Q}$. The matrix $\mathbf{Q}$ is called square root of matrix $\mathbf{P}$. Hence, from these facts, we have
\begin{align*}
	||\V_{n}^{-1/2}||_{F}^2 &= \Tr \left( \V_{n}^{-1/2} ( \V_{n}^{-1/2} )^{T}\right). \nonumber\\
                       &= \Tr \left( \V_{n}^{-1/2} \V_{n}^{-1/2} \right)\quad \left[\because V_{n}^{-1/2} \text{ is a symmetric matrix} \right].\nonumber\\
					   &= \Tr(\V_{n}^{-1}). \nonumber\\
					   & = \frac{\F_2+\G_2}{(\F_2\G_2-\langle\mathbf{f}, \mathbf{g}\rangle^2)}.\numberthis\label{eq:eqn30}
\end{align*}

%
%

\noindent We need to show 
$$\lim_{n \to \infty} ||{\V_n}^{-\frac{1}{2}}||_F^{2+\delta} \sum_{i=1}^{n} \E[||X_{i}||^{2+\delta}] = 0.$$ 

\begin{align*}
    &||{\V_n}^{-\frac{1}{2}}||_F^{2+\delta} \sum_{i=1}^{n} \E[||X_{i}||^{2+\delta}]\\ 
    & = \left(\frac{\F_2+\G_2}{(\F_2\G_2-\langle\mathbf{f}, \mathbf{g}\rangle^2)}\right)^{\frac{2+\delta}{2}} \sum_{i=1}^{n}((f_{i}^2 + g_{i}^2)^{\frac{2+\delta}{2}}). \numberthis\label{eq:eqn31}\\ 
     & =  \left(\frac{\F_2+\G_2}{(\F_2\G_2-\langle\mathbf{f}, \mathbf{g}\rangle^2)}\right)^{\frac{2+\delta}{2}} \sum_{i=1}^{n}((f_{i}^2 + g_{i}^2)^{\frac{2+\delta}{2}}).\\
    &=  \frac{1}{n^{\frac{\delta}{2}}} \left(\frac{ \frac{\F_2}{n} + \frac{\G_2}{n} }{\frac{\F_2}{n}\cdot \frac{\G_2}{n}- \frac{\langle\mathbf{f}, \mathbf{g}\rangle^2}{n^2}}\right)^{\frac{2+\delta}{2}} \sum_{i=1}^{n} \left(\frac{(f_{i}^2 + g_{i}^2)^{\frac{2+\delta}{2}}}{n}  \right). \\
    &=  \frac{1}{n^{\frac{\delta}{2}}} \times \left(\frac{ \sum_{i=1}^{n}\frac{f_{i}^2}{n} + \sum_{i=1}^{n} \frac{g_{i}^2}{n} }{\sum_{i=1}^{n}\frac{f_{i}^2}{n} \sum_{i=1}^{n}\frac{g_{i}^2}{n} - \left(\sum_{i=1}^{n} \frac{f_{i}g_{i}}{n}\right)^2}\right)^{\frac{2+\delta}{2}} \times\sum_{i=1}^{n} \left(\frac{(f_{i}^2 + g_{i}^2)^{\frac{2+\delta}{2}}}{n}  \right). \\
    &=  \frac{1}{n^{\frac{\delta}{2}}}  \times \left(\frac{ \E[f_{i}^2] + \E[g_{i}^2] }{\E[f_{i}^2]\cdot\E[g_{i}^2] - \left(\E[f_{i}g_{i}]\right)^2}\right)^{\frac{2+\delta}{2}} \times \E\left[(f_{i}^2 + g_{i}^2)^{\frac{2+\delta}{2}}\right].\\
    &  {\to 0 ~~~~~ \text{as}~~~~~ n \to \infty}. \numberthis\label{eq:eqn35}
\end{align*}
Equation~\eqref{eq:eqn31} holds due to Equation~\eqref{eq:eqn30} along with Equation~\eqref{eq:eqn22}. 
 {Finally, Equation~\eqref{eq:eqn35} holds due to Theorem~\ref{thm:multivariate_Lyapunov_CLT} because $0<\E\left[(f_{i}^2 + g_{i}^2)^{\frac{2+\delta}{2}}\right]<\infty $. }
Thus due to Theorem~\ref{thm:multivariate_Lyapunov_CLT}, we have 
\begin{align*}
 &\V_{n}^{-1/2}
\sum_{i=1}^n \X_{i}
\overset{d}{\to} \mathcal{N} \left( \mathbf{0}, \mathbf{I}
 \right). \\
 \implies
 &\V_{n}^{-1/2}
\sum_{i=1}^n \begin{bmatrix}
{f_i}{Y_i} \\
{g_i}{Y_i}
\end{bmatrix}
 = \V_{n}^{-1/2}\begin{bmatrix}
\tilde{f} \\
\tilde{g}
\end{bmatrix}
\overset{d}{\to}  \mathcal{N} \left( \mathbf{0}, \mathbf{I}
 \right). \numberthis\label{eq:eq600}
 \pushQED{\qed}
\end{align*}
\end{proof}


\subsubsection*{Proof of Corollary~\ref{cor:cor_bivariate}:}
\begin{proof}
    We know that for any random vector $\mathbf{X}$ of dimension $N\times1$ and any random matrix $\mathbf{A}$ of dimension $N\times N$
    \begin{align*}
        \E[\mathbf{A}\mathbf{X}] &= \mathbf{A}\cdot\E[\mathbf{X}].\\
        \Cov(\mathbf{A}\mathbf{X}) &= \mathbf{A} \cdot\Cov(\mathbf{X})\cdot \mathbf{A}^{T}.
    \end{align*}
    Therefore,
\begin{align*}
    \E\left[\V_{n}^{\frac{1}{2}} \V_{n}^{-1/2}\begin{bmatrix}
   \tilde{f} \\
    \tilde{g}
    \end{bmatrix}\right] &= \V_{n}^{\frac{1}{2}} \E\left[\V_{n}^{-1/2}\begin{bmatrix}
    \tilde{f} \\
    \tilde{g}
    \end{bmatrix} \right].\\
    &= \V_{n}^{\frac{1}{2}} \mathbf{0} = \mathbf{0}.\\
    \implies \E\left[\V_{n}^{\frac{1}{2}} \V_{n}^{-1/2}\begin{bmatrix}
        \tilde{f} \\
    \tilde{g}
    \end{bmatrix}\right] &= \E\begin{bmatrix}
        \tilde{f} \\
    \tilde{g}
    \end{bmatrix} = \mathbf{0}. \numberthis \label{eq:cor_1}
\end{align*}
\begin{align*}
    \Cov\left(\V_{n}^{\frac{1}{2}} \V_{n}^{-1/2}\begin{bmatrix}
    \tilde{f} \\
    \tilde{g}
    \end{bmatrix} \right) &= \V_{n}^{\frac{1}{2}}\Cov \left(\V_{n}^{-1/2}\begin{bmatrix}
    \tilde{f} \\
    \tilde{g}
    \end{bmatrix} \right) \left(\V_{n}^{\frac{1}{2}}\right)^{T}. \\
    &= \V_{n}^{\frac{1}{2}} \mathbf{I} \left(\V_{n}^{\frac{1}{2}}\right)^{T}.\\ 
    &= \V_{n}^{\frac{1}{2}} \left(\V_{n}^{\frac{1}{2}}\right)^{T}.\\
    &= \V_{n}.\\
    \implies\Cov\left(\begin{bmatrix}
    \tilde{f} \\
    \tilde{g}
    \end{bmatrix} \right) &= \V_{n}.  \numberthis \label{eq: cor_2}
\end{align*}

Equation \ref{eq:cor_1} and \ref{eq: cor_2} implies
\begin{align*}
   \begin{bmatrix}
\tilde{f} \\
    \tilde{g}
\end{bmatrix}
&\overset{d}{\to} \mathcal{N} \left( \mathbf{0},
\mathbf{V}_{n} \right). 
\end{align*}
\end{proof}

\subsection{Some basic facts from probability theory:}

We use a few facts from probability theory for our analysis purpose. 
\begin{newfact}\label{fact:prob}
Let $X,Y,X_i,$ and $Y_i$   are  the random variables and $a, b, a_i,$ and $ b_i$   are the constants. Then
\begin{itemize}
\item $\E[aX]=a\E[X].$
\item $\Var[aX]=a^2\Var[X]$ and $\Var[a+X]=\Var[X]$.
    \item $\Var \left( \sum_{i\in [n]}X_i \right)=\sum_{i\in [n]}\Var \left( X_i \right)+\sum_{i\neq j, i,j\in [n]}\Cov[X_i, X_j].$
    \item $\Cov[aX, Y]=a\Cov[X, Y]$ and $\Cov[a+X, Y]=\Cov[X, Y]$.
    \item $\Cov\left[\sum_{i\in [n]}a_iX_i, \sum_{i\in [m]}b_iY_i \right]=\sum_{i\in [n]}\sum_{j\in [m]} a_ib_j\Cov\left[X_i, Y_j \right].$
\end{itemize}
\end{newfact}


\end{document}